\documentclass[twocolumn]{bmcart}

\usepackage{amsthm,amsmath}
\usepackage{bm}
\usepackage{hyperref}
\usepackage{color}
\usepackage{graphicx}
\usepackage[utf8]{inputenc}
\usepackage{flushend}

\usepackage{layout}
\usepackage{wrapfig}
\usepackage{txfonts}

\usepackage{url}
\usepackage{algorithm}
\usepackage{algcompatible}
\usepackage{algpseudocode}
\pdfoutput=1 

\newcommand{\tabref}[1]{{Table.\ref{#1}}}
\newcommand{\figref}[1]{{Fig.\ref{#1}}}

\newcommand{\argmax}{\mathop{\mathrm{argmax}}\limits}

\def\introduction{Introduction}
\def\method{Method of human mimetic environmental sound recognition system}
\def\experiment{Experimental results}
\def\discussion{Discussion}
\def\conclusion{Conclusion}

\startlocaldefs
\endlocaldefs

\begin{document}

\begin{frontmatter}

\begin{fmbox}
\dochead{Research Article}

\title{
  Human-mimetic binaural ear design and sound source direction estimation for task realization of musculoskeletal humanoids
}

\author[
  addressref={aff1},
  email={oomura@jsk.imi.i.u-tokyo.ac.jp}
]{\inits{Y.O.}\fnm{Yusuke} \snm{Omura}}
\author[
  addressref={aff1},
  email={kawaharazuka@jsk.imi.i.u-tokyo.ac.jp}
]{\inits{K.K.}\fnm{Kento} \snm{Kawaharazuka}}
\author[
  addressref={aff1},
  email={nagamatsu@jsk.imi.i.u-tokyo.ac.jp}
]{\inits{Y.N.}\fnm{Yuya} \snm{Nagamatsu}}
\author[
  addressref={aff1},
  email={koga@jsk.imi.i.u-tokyo.ac.jp}
]{\inits{Y.K.}\fnm{Yuya} \snm{Koga}}
\author[
  addressref={aff1},
  email={nishiura@jsk.imi.i.u-tokyo.ac.jp}
]{\inits{M.N.}\fnm{Manabu} \snm{Nishiura}}
\author[
  addressref={aff1},
  email={toshimitsu@jsk.imi.i.u-tokyo.ac.jp}
]{\inits{Y.T.}\fnm{Yasunori} \snm{Toshimitsu}}
\author[
  addressref={aff1},
  email={asano@jsk.imi.i.u-tokyo.ac.jp}
]{\inits{Y.A.}\fnm{Yuki} \snm{Asano}}
\author[
  addressref={aff1},
  email={k-okada@jsk.imi.i.u-tokyo.ac.jp}
]{\inits{K.O.}\fnm{Kei} \snm{Okada}}
\author[
  addressref={aff2},
  email={koji_kawasaki@mail.toyota.co.jp}
]{\inits{K.K.}\fnm{Koji} \snm{Kawasaki}}
\author[
  addressref={aff1},
  email={inaba@jsk.imi.i.u-tokyo.ac.jp}
]{\inits{M.I.}\fnm{Masayuki} \snm{Inaba}}

\address[id=aff1]{
  \orgdiv{Department of Information Science and Technology},
  \orgname{The University of Tokyo},
  \city{Tokyo},
  \cny{Japan},
}
\address[id=aff2]{
  \orgname{TOYOTA MOTOR CORPORATION}
}

\begin{abstractbox}

\begin{abstract}

Human-like environment recognition by musculoskeletal humanoids is important for task realization in real complex environments and for use as dummies for test subjects.
Humans integrate various sensory information to perceive their surroundings, and hearing is particularly useful for recognizing objects out of view or out of touch.
In this research, we aim to realize human-like auditory environmental recognition and task realization for musculoskeletal humanoids by equipping them with a human-like auditory processing system.
Humans realize sound-based environmental recognition by estimating directions of the sound sources and detecting environmental sounds based on changes in the time and frequency domain of incoming sounds and the integration of auditory information in the central nervous system.
We propose a human mimetic auditory information processing system, which consists of three components: the human mimetic binaural ear unit, which mimics human ear structure and characteristics, the sound source direction estimation system, and the environmental sound detection system, which mimics processing in the central nervous system.
We apply it to Musashi, a human mimetic musculoskeletal humanoid, and have it perform tasks that require sound information outside of view in real noisy environments to confirm the usefulness of the proposed methods.

\end{abstract}


\end{abstractbox}

\end{fmbox}

\end{frontmatter}

\section*{\introduction} \label{sec:introduction}
The musculoskeletal humanoids \cite{nakanishi2013design}, \cite{wittmeier2013toward}, \cite{jantsch2013anthrob}, \cite{asano2016kengoro}, which mimic the human body structure in detail, are expected to be used for environmental contact behavior and as subject dummies by using their flexible body structure effectively.
In order to use them as substitutes for humans, they need to recognize environments as humans do.
Research on vision using movable eyes \cite{makabe2018eyeunit} and tactile sense using the hand \cite{makino2018hand} and foot \cite{shinjo2019foot} with distributed force sensors have been conducted.
However, humans use not only visual and tactile senses but also auditory perception to recognize the state of objects out of view or out of touch.
In order to have musculoskeletal humanoids perform tasks based on environmental recognition and use them as subject dummies, human mimetic auditory information processing is also essential in addition to vision and tactile senses.
There are two major types of information that we can obtain through hearing: sound source directions and sound types.
Humans realize sound-based environmental recognition by integrating these pieces of information.

As for existing sound source direction estimation approaches, high-resolution methods, such as the MUSIC (MUltiple SIgnal Classification) method \cite{MUSIC}, have been proposed, and research using robot audition have been conducted widely.
However, most methods are proposed for multi-channel systems with three or more channels, and in the case of human-like inputs with only two channels, the accuracy of sound source direction estimation in 3D is reduced due to the influence of background noise in the real environment.
GCC-PHAT (Generalized Cross-Correlation PHAse Transform) \cite{GCC-PHAT}, which targets two channels of input, uses the time difference between left and right inputs to estimate the sound source direction, but the range of estimated direction is limited.
Also, in order to improve the accuracy of those methods, the required sound sampling length becomes longer.

In robot audition, which aims to achieve human-like environmental recognition, environmental sound recognition using a pair of microphones is a particularly important issue.
Sound source localization and understanding the meaning of sound in various scenes are important in task realization, and integrating them with other senses enables more accurate environmental recognition.
SIG \cite{nakadai2000active}, a humanoid that has a human-like head and torso, achieves tracking while filtering sound inputs based on epipolar geometry and vision.
The sound recognition system, which integrates sound source localization, sound source separation, and speech recognition \cite{nakadai2004improvement}, achieves simultaneous recognition of multiple speech signals by filters depending on sound source directions.
Also, implementation of this method into a robot in the real world is conducted by using a reconfigurable processor \cite{kurotaki2005implementation}.

It is widely acknowledged that human sound source direction estimation is related to the complex unevenness in pinnae on the incoming sound.
Therefore, research focusing on the shape of the human outer ear has been conducted.
In sound source direction estimation using a robot with reflectors shaped like pinnae and a monocular camera \cite{nakashima20053d}, the changes in frequency response caused by the reflectors are learned using a self-organizing map.
3D sound source direction estimation is conducted, but it is unsuitable for real environment applications because it requires a single sound source and a broadband signal.
The sound source direction estimation method with weighting based on signal-to-noise ratio in addition to GCC-PHAT \cite{Kim2015improved} is applied to SIG-2 \cite{sig2}, a humanoid with human mimetic outer ears.
It is possible to estimate the sound source direction for multiple sound sources in the real environment, but there is still the problem that the estimated direction is limited.
Sound source localization methods inspired by the human central nervous system \cite{heckmann2006auditory}, \cite{youssef2013learning}, \cite{davila2018enhanced} learn the relationship between the sound source direction and the difference of the left and right sounds.
It is possible to estimate the sound source direction in the echoic and noisy environment, but the estimated direction is limited to the front of the horizontal plane.
A two-channel sound source direction estimation method that can be used in the real noisy environment is not limited to broadband sounds, and has a wide range of estimated directions, must be researched.

On the other hand, a number of neural network-based methods for environmental sound detection have been proposed.
EnvNet \cite{tokozume2017learning} uses sound waves as input to a convolutional neural network (CNN), and acoustic feature maps are formed in the CNN, resulting in highly accurate environmental sound recognition.
In addition, the convolutional recurrent neural network-based method \cite{iqbal2018stacked} realized environmental sound detection by using a model that captures temporal changes in sound.
Although these methods achieve high accuracy in detecting environmental sound, they are not designed for real-time use, so the required sound sampling length becomes longer, and they are still not suitable for real-time operation.

\begin{figure}[tb]
  \centering
  \includegraphics[width=0.9\columnwidth]{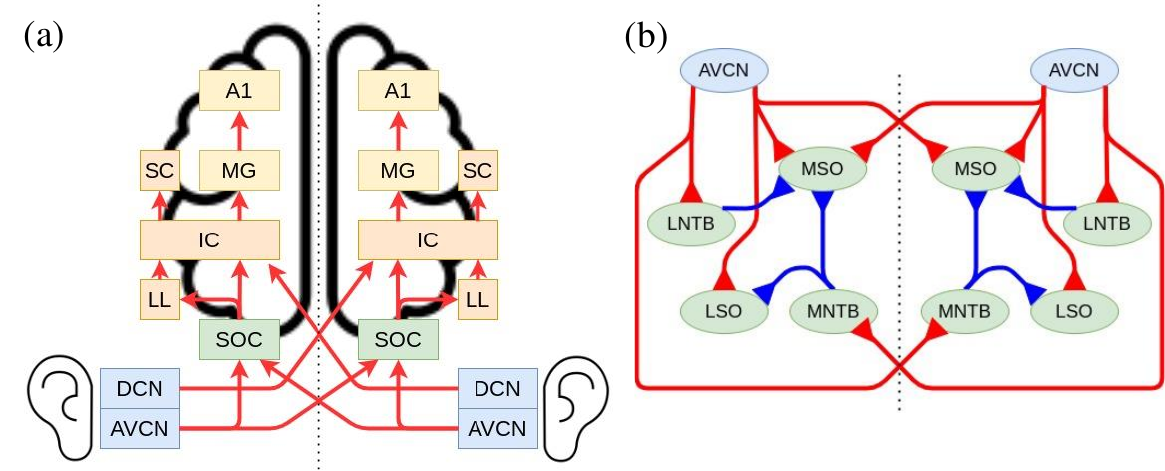}
  \caption{
    (a) Pathways of acoustic information in mammals.
    (b) ILD and ITD detection circuit in SOC.
    Red wires mean excitatory projection. 
    Blue wires mean inhibitory projection.
    MSOs detect ITD and LSOs detect ILD.
  }
  \label{fig:human-ssde}
\end{figure}

\begin{figure}[tb]
  \centering
  \includegraphics[width=0.9\columnwidth]{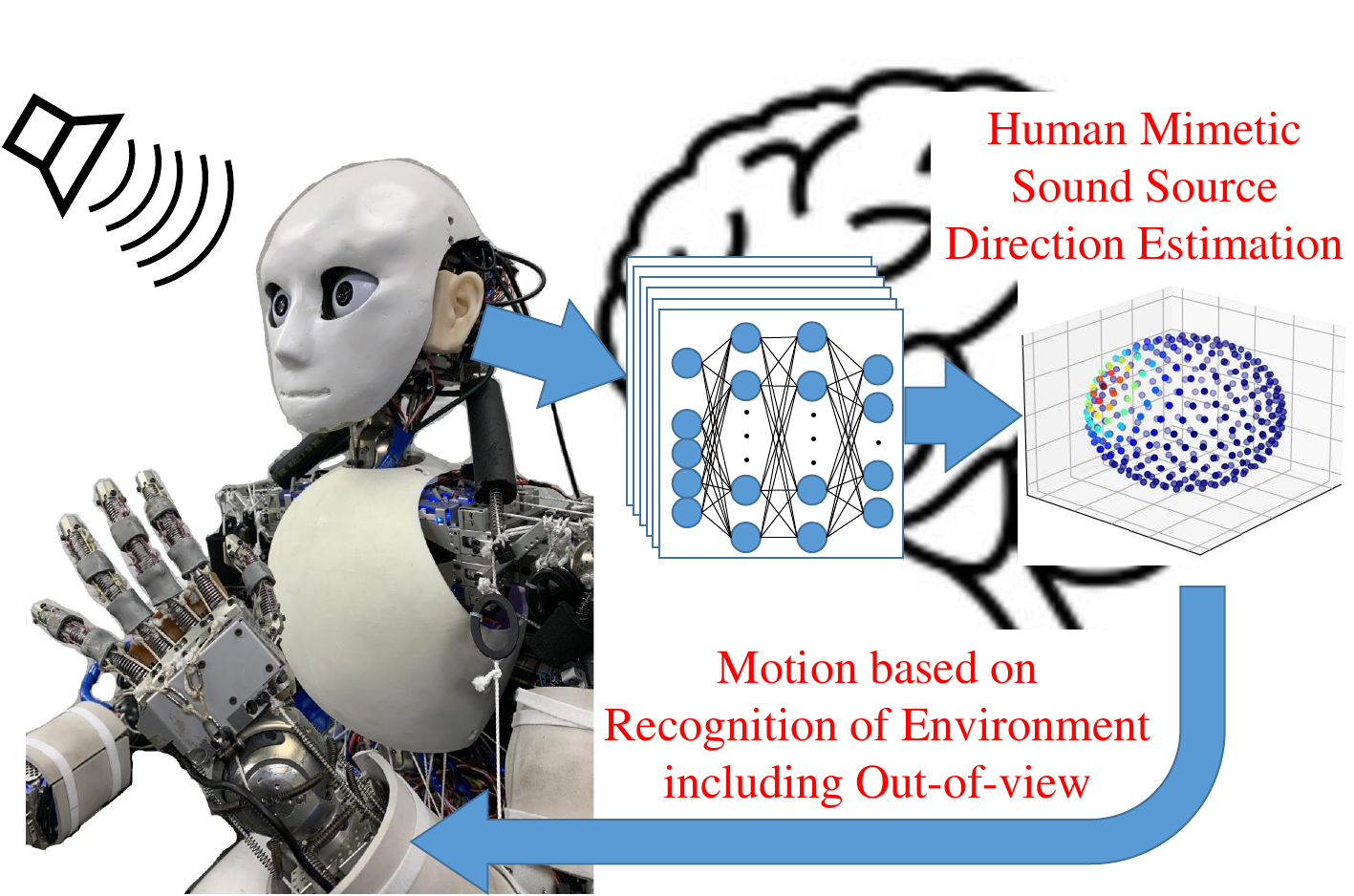}
  \caption{The concept of this study}
  \label{fig:concept}
\end{figure}

The human ear is divided into three parts based on its structural and functional characteristics, the outer ear, middle ear, and inner ear.
The outer ear has a pinna with complex unevenness and shows complex frequency response with a sharp gain reduction notch depending on the sound source direction \cite{musicant1984influence}.
Also, sound source direction estimation accuracy is significantly decreased when pinna unevenness is filled \cite{gardner1973problem}.
In robot audition, implementation of pinnae in a telepresence robot increases sound source localization accuracy of robot users in the median plane \cite{toshima2009possibility}.
Also, this study says implementing pinnae and head movement increase the sound source localization of telepresence robot users.
This complex relationship can be investigated by measuring the Head-Related Transfer Function (HRTF), which describes the relationship between the source sound signal and the actual incoming sound.

Auditory information processing in the central nervous system of mammals is shown in \figref{fig:human-ssde}.
Auditory information decomposed into frequency components in the cochlea is transmitted to the cochlear nucleus (CN).
Signals are transmitted from the anterior ventral cochlear nucleus (AVCN) to the superior olivary complex (SOC), the interaural level difference (ILD), and the interaural time difference (ITD) are extracted in SOC.
The lateral nucleus of trapezoid body (LNTB) transmits the contralateral signal as inhibitory projection, and the medial nucleus of trapezoid body (MNTB) transmits the ipsilateral signal as inhibitory projection.
ITD is obtained by detecting the simultaneity of sound when the medial superior olive (MSO) receives projections from each side.
ILD is obtained by detecting the firing frequency changes by comparing the intensity difference when the lateral superior olive (LSO) receives excitatory projections from the ipsilateral side, and inhibitory projections from the contralateral side \cite{grothe2010mechanisms}.
ILD and ITD extracted in SOC are transmitted to the lateral lemniscus (LL) and the inferior colliculus (IC) and integrated.
It is confirmed that there are neurons in IC that respond to sound from specific directions \cite{semple1983spatial}.
Auditory information is transmitted from IC to the medial geniculate body (MG) and the primary auditory cortex (A1).
Also, auditory information from IC is transmitted to the superior colliculus (SC), which mainly processes visual information, and the spatial map is formed in SC \cite{palmer1982representation}.

It can be summarized that the environmental sound recognition function is composed of the following functions;
\begin{enumerate}
  \item physical effects on frequency response by the complex unevenness of the pinna and frequency decomposition of incoming sounds
  \item sound source direction estimation based on detection of ILD and ITD, and neurons which respond to each sound source directions
  \item environmental sound detection based on the integration of frequency components and temporal changes of incoming sounds.
\end{enumerate}

In this research, we propose a human mimetic auditory information processing system composed of a human mimetic binaural ear, sound source direction estimation system, and environmental sound detection system.
The concept of this study is shown in \figref{fig:concept}.
The rest of this paper is organized as follows.
In "\method" section, we explain the proposed system and methods of sound source direction estimation and environmental sound detection.
In "\experiment" section,  we apply the proposed method to a musculoskeletal humanoid and conduct task realization experiments that require information out of view.
Finally, the discussion and conclusion are presented.

\section*{\method} \label{sec:method}

\begin{figure}[tb]
  \centering
  \includegraphics[width=0.9\columnwidth]{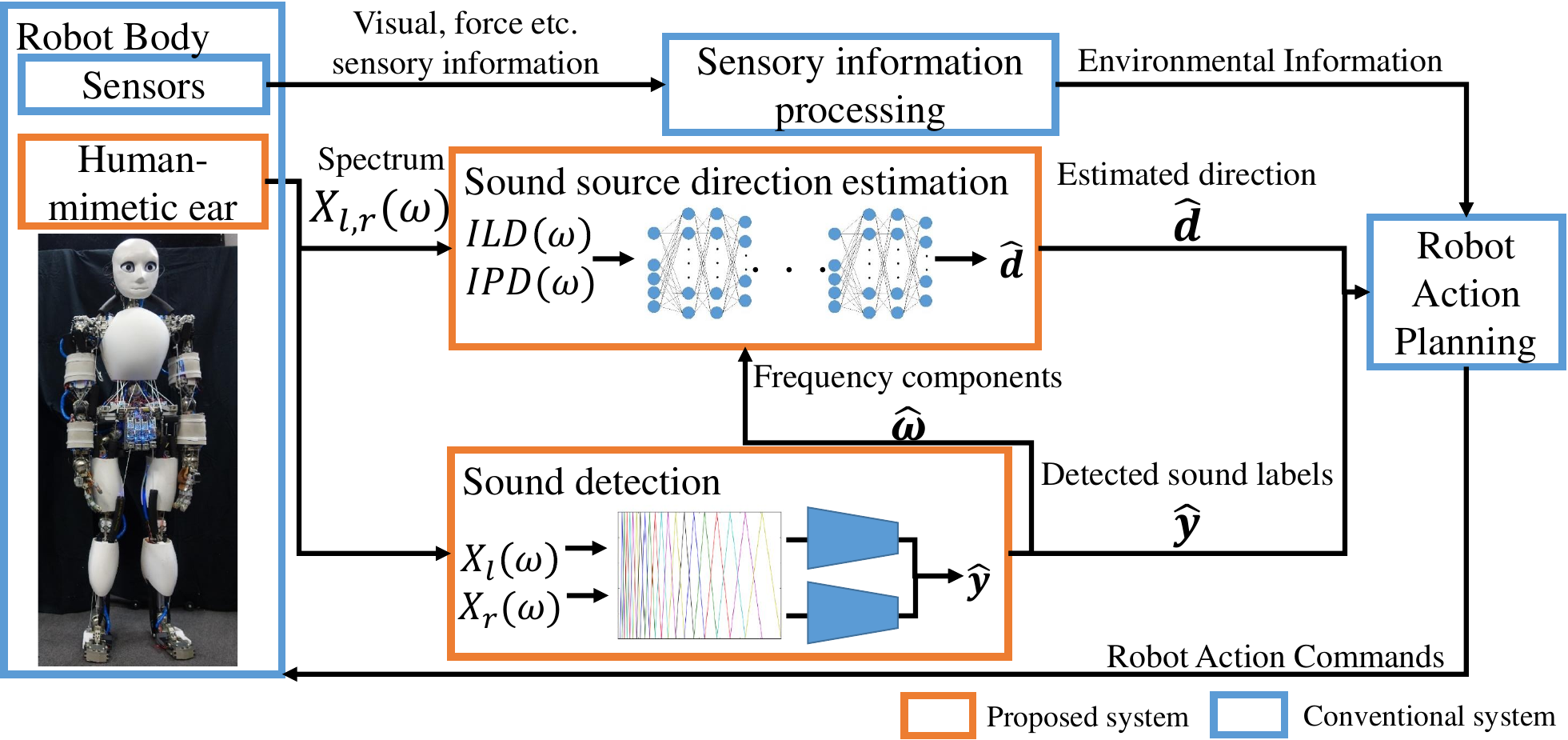}
  \caption{Overview of proposed system for Human mimetic environmental sound recognition system.}
  \label{fig:whole-system}
\end{figure}

We propose an auditory information processing system as shown in \figref{fig:whole-system}.
We explain the components of this system in the following subsections.

\subsection*{Human Mimetic Binaural Ear}

\begin{figure}[tb]
  \centering
  \includegraphics[width=0.9\columnwidth]{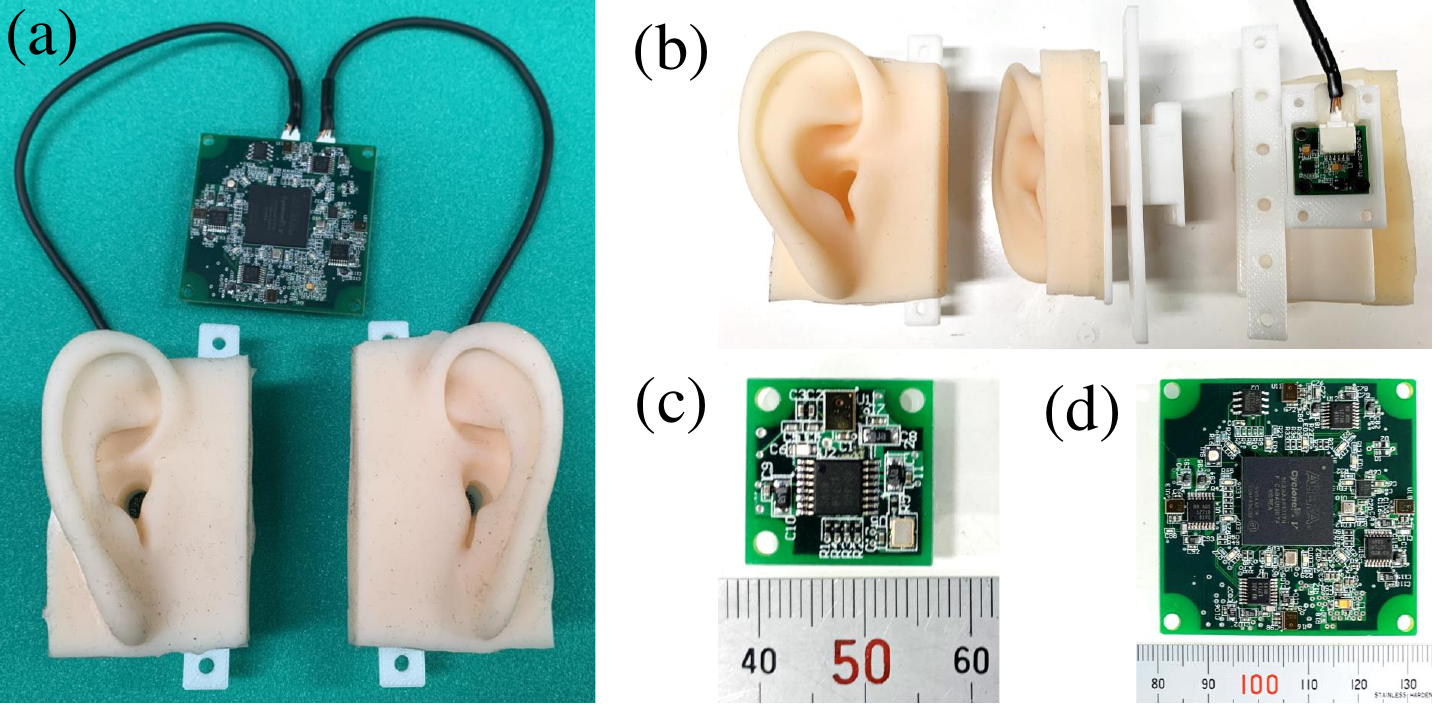}
  \caption{(a) Overview of developed human mimetic binaural ear unit.
  (b) Human mimetic outer ear structure.
  (c) Microphone board.
  (d) Acoustic processing board.}
  \label{fig:ear}
\end{figure}

In this research, we develop the human mimetic binaural ear unit as shown in \figref{fig:ear} in order to realize the functions (i) and (ii).
It is composed of a human mimetic outer ear structure, microphone board, and acoustic processing board and transmits incoming sound information to the humanoid internal computer.

Its pinna and cartilaginous part of the ear canal is made of silicone rubber, and its bony part of the ear canal and container of microphone board is made of 3D printed ABS resin.
The human mimetic outer ear structure has a pinna and ear canal.
The distance between the microphone and the open end of the ear canal is approximately 26 mm, and it is about the same length as the human ear canal.
The microphone only measures the sound passing its ear canal by covering it with silicone rubber.

The microphone board has a MEMS microphone (ICS-40619) and a 24-bit ADC (ADS1271B) and has human-like acoustic characteristics, as shown in \tabref{tab:mic}.
This MEMS microphone has a flat frequency response, particularly from 100 Hz to 10000 Hz, covering everyday sounds and speech.
The sampling rate is 44100 Hz.

\begin{table}[tb]
  \centering
  \caption{Comparison between the developed microphone board and the human ear.}
  \label{tab:mic}
  \begin{tabular}{ccc}
    \hline
    & Mic. board & Human ear \\ \hline
    Frequency Response[Hz] & 20 $\sim$ 20000 & 20 $\sim$ 20000\\ \hline
    Absolute Threshold[$\mu$Pa] & 13.4 & $\sim$ 20 \\ \hline
    Acoustic Overload Point[dBSPL] & 132 & $\sim$ 130 \\ \hline
    Sensitivity[dBA] & -39 ~ -37 & - \\ \hline
    Signal-to-Noise Ratio[dBA] & 67 & - \\ \hline
  \end{tabular}
\end{table}

The acoustic processing board calculates spectra from incoming sounds by FPGA (Cyclone V).
Each spectrum $X_{l,r}$ can be calculated as follows,

\begin{gather}
  x_{l,r}^{\prime}(t) = w_{ham}(t) x_{l,r}(t) \\
  X_{l,r}(\omega) = \mathrm{FFT}[x_{l,r}^{\prime}(t)] \label{eqn:fft} \\
  t = 0,\cdots,N-1,\:\:\:\:\:\:\: \omega = 0,\cdots,N-1
\end{gather}

where $x_{l,r}$ is each incoming sound signal, $x_{l,r}^{\prime}$ is each incoming sound signal after windowing, $w_{ham}$ is a hamming window, $*$ is convolution, $\mathrm{FFT}$ is fast Fourier transform, $t$ is time, $\omega$ is discrete frequency, and $N$ is the number of FFT points, 2048 in this study.
Each spectrum calculated by FFT is transmitted to the humanoid internal computer in a cycle of 62.5 Hz.

\subsection*{Human Mimetic Sound Source Direction Estimation}

Sound source direction estimation described as (ii) is conducted based on left and right spectra.
We propose a neural network-based sound source direction estimation method that mimics the process of the human central nervous system.
ILD and ITD are projected to IC in humans, but we use interaural phase difference (IPD) instead of ITD in this study.
ILD and IPD, inputs for sound source direction estimation, are calculated based on left and right spectra.
ILD and IPD can be calculated as follows,

\begin{gather}
  \hat{X}_{l,r}(\omega_{i}) = \frac{X_{l,r}(\omega_{i})}{\sqrt{|X_{l}(\omega_{i})|^{2}+|X_{r}(\omega_{i})|^{2}}}\\
  ILD(\omega_{i}) = \log{|\hat{X}_{l}(\omega_{i})|} - \log{|\hat{X}_{r}(\omega_{i})|}\\
  IPD(\omega_{i}) = 
  \begin{pmatrix}
    \mathrm{Real}[\hat{X}_{l}(\omega_{i})/|\hat{X}_{l}(\omega_{i})|]\\
    \mathrm{Imag}[\hat{X}_{l}(\omega_{i})/|\hat{X}_{l}(\omega_{i})|]\\
    \mathrm{Real}[\hat{X}_{r}(\omega_{i})/|\hat{X}_{r}(\omega_{i})|]\\
    \mathrm{Imag}[\hat{X}_{r}(\omega_{i})/|\hat{X}_{r}(\omega_{i})|]
  \end{pmatrix}
\end{gather}

where $\hat{X}_{l,r}$ is left and right normalized spectra at discrete frequency $\omega_{i}$, $\mathrm{Real}$ is the operation to take out the real part, and $\mathrm{Imag}$ is the operation to take out the imaginary part.
In order to use the spectra obtained by FFT as the input of the neural network with as little modification as possible, the ILD and IPD are determined by trial and error. 

\begin{figure}[tb]
  \centering
  \includegraphics[width=0.9\columnwidth]{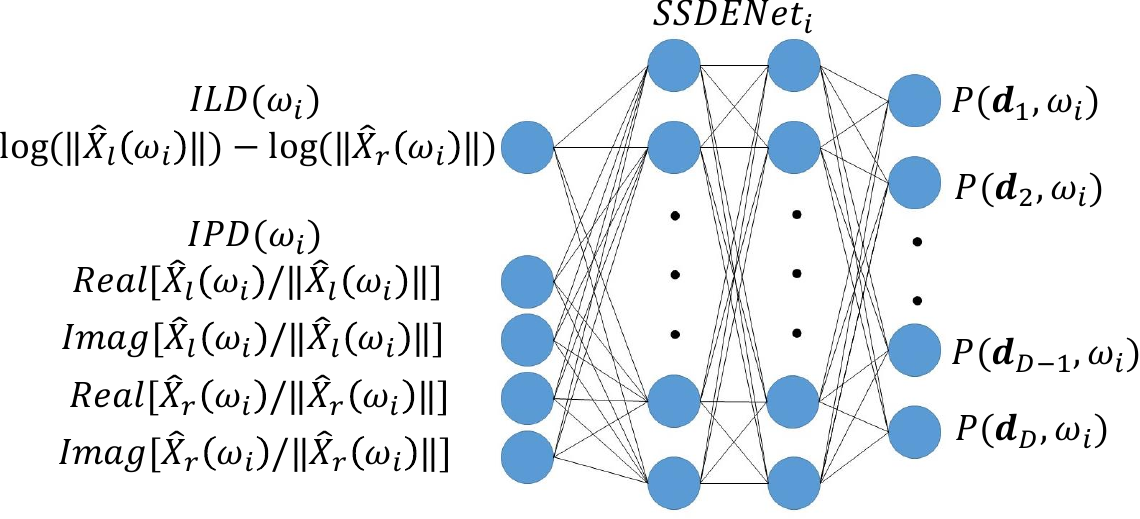}
  \caption{Structure of SSDENet.}
  \label{fig:ssdenet}
\end{figure}

For each frequency, sound source existences $\bm{P(\omega_{i})}$ can be calculated by Sound Source Direction Estimation Network (SSDENet) shown in \figref{fig:ssdenet}, using calculated ILD and IPD as input, as follows,

\begin{gather}
  \bm{P}(\omega_{i}) = \mathrm{SSDENet}_{i}(ILD(\omega_{i}), IPD(\omega_{i}))\\
  \bm{P}(\omega_{i}) = 
  \begin{pmatrix}
    P(\bm{d}_{1}, \omega_{i}) & ... & P(\bm{d}_{D}, \omega_{i})
  \end{pmatrix}^{T}
\end{gather}

where $\mathrm{SSDENet}_{i}$ is SSDENet at discrete frequency $\omega_{i}$, $\bm{d}_{k}$ is a vector which expresses 3-dimensional sound source direction, $P(\bm{d}_{k}, \omega_{i})$ is a sound source existence at direction $\bm{d}_{k}$ and discrete frequency $\omega_{i}$, and $D$ is the number of directions $\bm{d}_{k}$ at which estimation is conducted.
SSDENet is a neural network consisting of 4 fully connected layers, with the number of nodes in each layer being 5, 500, 500, and 326, and the activation function is sigmoid.
Since SSDENet outputs sound source existences of all directions at once and is structured so that the sound source directions affect each other, it is expected to reduce false estimates more than existing engineering methods that estimate each direction independently.

Estimated sound source direction $\hat{\bm{d}}$ can be calculated as follows,

\begin{gather}
  \hat{\bm{d}} = \argmax_{\bm{d}_{k}} P(\bm{d}_{k}) \text{  } (k=1,\dots,D)\\
  P(\bm{d}_{k}) = \sum_{i} c_{i}P(\bm{d}_{k}, \omega_{i})\\
  c_{i} = 
  \begin{cases}
    1 & (\text{if $\omega_{i}$ is valid})\\
    0 & (\mathrm{otherwise})\\
  \end{cases}
\end{gather}

where $c_{i}$ is a coefficient indicating that the discrete frequency $\omega_{i}$ is valid for estimation.
$\omega_{i}$ can be set to match each sound to be estimated.
The sound source direction can be estimated while excluding unnecessary frequency components of the subject sound by analytically selecting the frequency components.
By estimating only the frequencies of the sound source signal, the influence of other sounds can be suppressed, and the directions of multiple sound sources with different frequency components can be estimated.
In practice, the sound source direction estimation is conducted on the frequency components $\hat{\bm{\omega}}$ output by the binaural environmental sound detection described below.

In order to train SSDENets, we use pre-measured HRTFs and generate training data by placing virtual sound sources.
When a virtual sound source is placed in direction $\bm{d}_{h}$, the spectra $X_{l,r}^{train}$ and the sound source existence $P^{train}$ are calculated as follows,

\begin{gather}
  X_{l,r}^{train}(\omega_{i}) = \frac{A_{l,r}(\bm{d}_{h}, \omega_{i})}{\sqrt{|A_{l}(\bm{d}_{h}, \omega_{i})|^{2}+|A_{r}(\bm{d}_{h}, \omega_{i})|^{2}}}s+n_{l,r}\\
  P^{train}(\bm{d}_{k},\omega_{i}) = \frac{1}{2\pi\sigma^{2}}\exp(-\frac{\Delta d_{k,h}^{2}}{2\sigma^{2}})
\end{gather}

where $A_{l,r}$ are left and right HRTF, $s$ is spectrum of virtual sound source, $n_{l,r}$ are left and right background noise spectra, $\Delta d_{k,h}$ is the angle between the direction $\bm{d}_{k}$ and the direction $\bm{d}_{h}$, and $\sigma$ is the variance of the existence distribution.
By learning the sound source existences as a distribution with a peak in the correct sound source direction, it is expected for sound source direction to influence the close directions.
The loss function of learning is the mean squared error, and the learning rule is Adam.

\subsection*{Binaural Environmental Sound Detection}

\begin{figure}[tb]
  \centering
  \includegraphics[width=0.9\columnwidth]{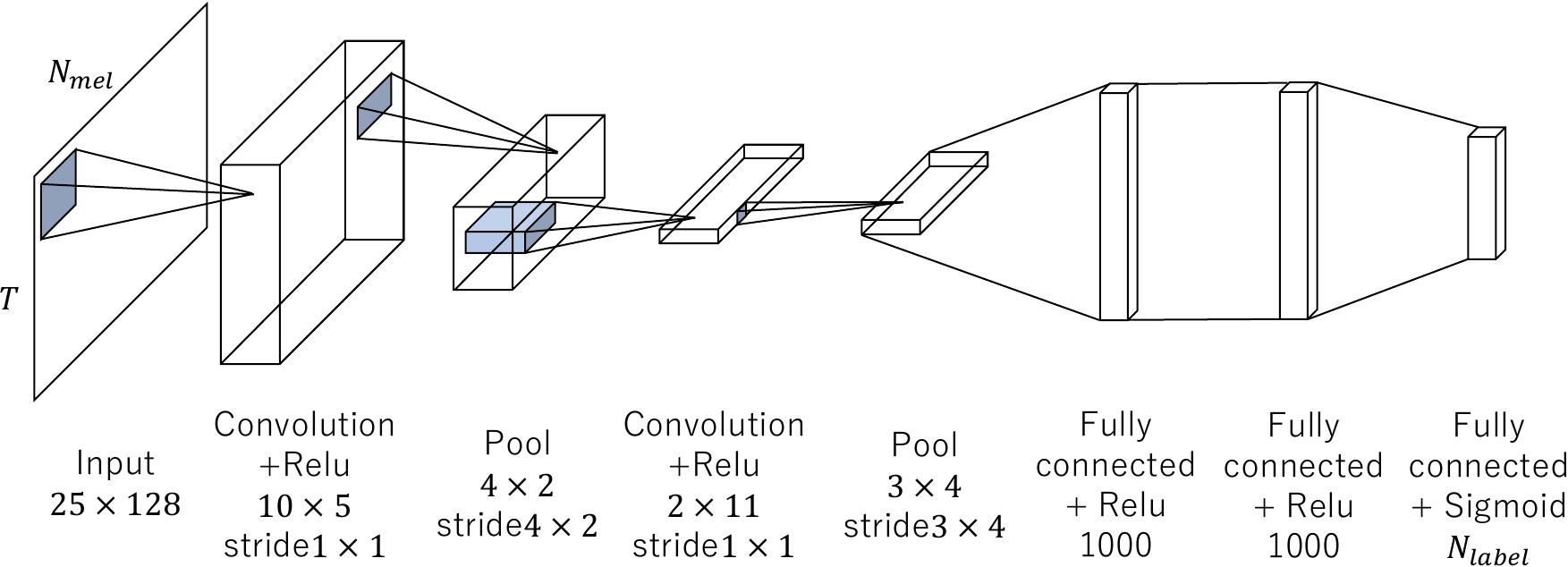}
  \caption{Structure of MelCNN based on log-MelCNN \cite{piczak2015environmental}.}
  \label{fig:melcnn}
\end{figure}

Environmental sound detection described as (iii) is conducted based on the convolutional neural network (MelCNN) using each Mel spectrogram as input.
The structure of MelCNN is based on logMel-CNN \cite{piczak2015environmental}, as shown in \figref{fig:melcnn}, and consists of two convolutional layers and two fully-connected layers.
LogMel-CNN is a simple neural network used in environmental sound recognition.
The input of logMel-CNN is a Mel spectrogram, and it can be easily modified in both time and frequency domains.
We think that logMel-CNN can be used as a detector by modified to target short sounds and running periodically in practice.
MelCNN uses a Mel spectrogram as input and outputs the existence probability of each sound by using sigmoid as the activation function of the output layer.
The Mel spectrogram is a series of 25 Mel spectra with 128 points on the mel-scale.
This Mel spectrogram is normalized to use relative intensity as the input because the volume of incoming sound can easily vary depending on the distance of the sound source in the real environment.
The existence probability of the target sound is defined as 0 when the sound is absent and as 1 when the sound is present.
MelCNN outputs the existence probabilities of each sound respectively.
In order to predict co-occurring labeled sounds, we use each sound data and mixed sound data for the training of MelCNN.
In case multiple labeled sounds co-occur when MelCNN is used in a real environment, the outputs of MelCNN are existences probabilities of each sound, and multiple sound detection can be expected.
Left and Right Mel spectrograms are input to MelCNN, respectively.
The existence probabilities of each sound are calculated by summing the outputs weighted by sound volume ratio as follows,

\begin{gather}
  p(y_{i}) = \frac{V_{l}p_{l}(y_{i}) + V_{r}p_{r}(y_{i})}{V_{l}+V_{r}}
\end{gather}

where $y_{i}$ is the label of sound, $p(y_{i})$ is the existence probability of sound $y_{i}$, $p_{l,r}(y_{i})$ is the left and right existence probabilities of sound $y_{i}$, and $V_{l,r}$ are the left and right sound volume.
This weighting is intended to prioritize the result of a more audible side.
The loss function of learning is the KL divergence, and the learning rule is Adam.

Also, the characteristic frequency components $\hat{\bm{\omega}}$ of each sound are calculated during train data processing.
When MelCNN predicts, detected sound labels and corresponding frequency components $\hat{\bm{\omega}}$ are output, and the frequencies are used for determining which frequency is used for sound source direction estimation.

\section*{\experiment} \label{sec:experiment}

\subsection*{Effects of Human Mimetic Outer Ear Structure}
\subsubsection*{Change of frequency response in the median plane}

\begin{figure}[tb]
  \centering
  \includegraphics[width=0.9\columnwidth]{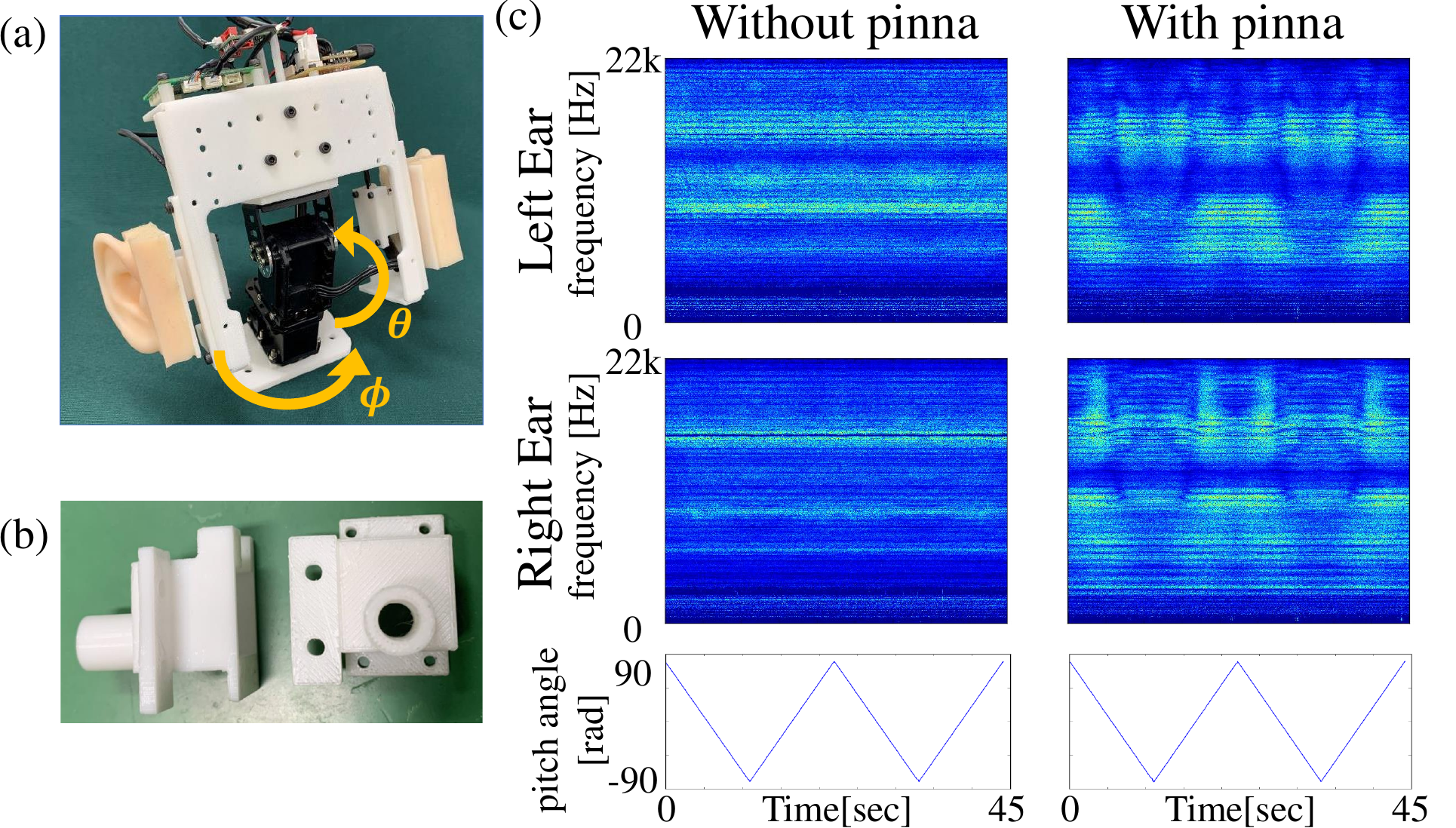}
  \caption{(a) Dummy head with developed ear unit.
  (b) Outer ear structure without pinna.
  (c) The difference of frequency response with and without pinna.}
  \label{fig:dummy}
\end{figure}

We investigate whether the pinnae of the human mimetic binaural ear unit affect the frequency response in the median plane.
A dummy head, as shown in \figref{fig:dummy}(a), is used in the experiment.
The dummy head can rotate in azimuth angle $\phi$ and elevation angle $\theta$ by two servo motors.

A loudspeaker is placed 1.5 m in front of the dummy head to play white noise, while the elevation angle is varied from $-90^{\circ}$ to $90^{\circ}$.
In the experiment, we compare the outer ear structure with and without pinnae, as shown in \figref{fig:dummy}(b).
The results of this experiment are shown in \figref{fig:dummy}(c).
While moving the neck in elevation, there is little change in the frequency response in the absence of pinnae.
However, in the case of outer ear structure with pinnae, the frequency response changes significantly as the elevation changes.
In particular, notches occur clearly in the high-frequency band above approximately 10000 Hz, suggesting that introducing the human mimetic outer ear structure can induce notches like humans.

\subsubsection*{HRTF measurement}

\begin{figure}[tb]
  \centering
  \includegraphics[width=0.9\columnwidth]{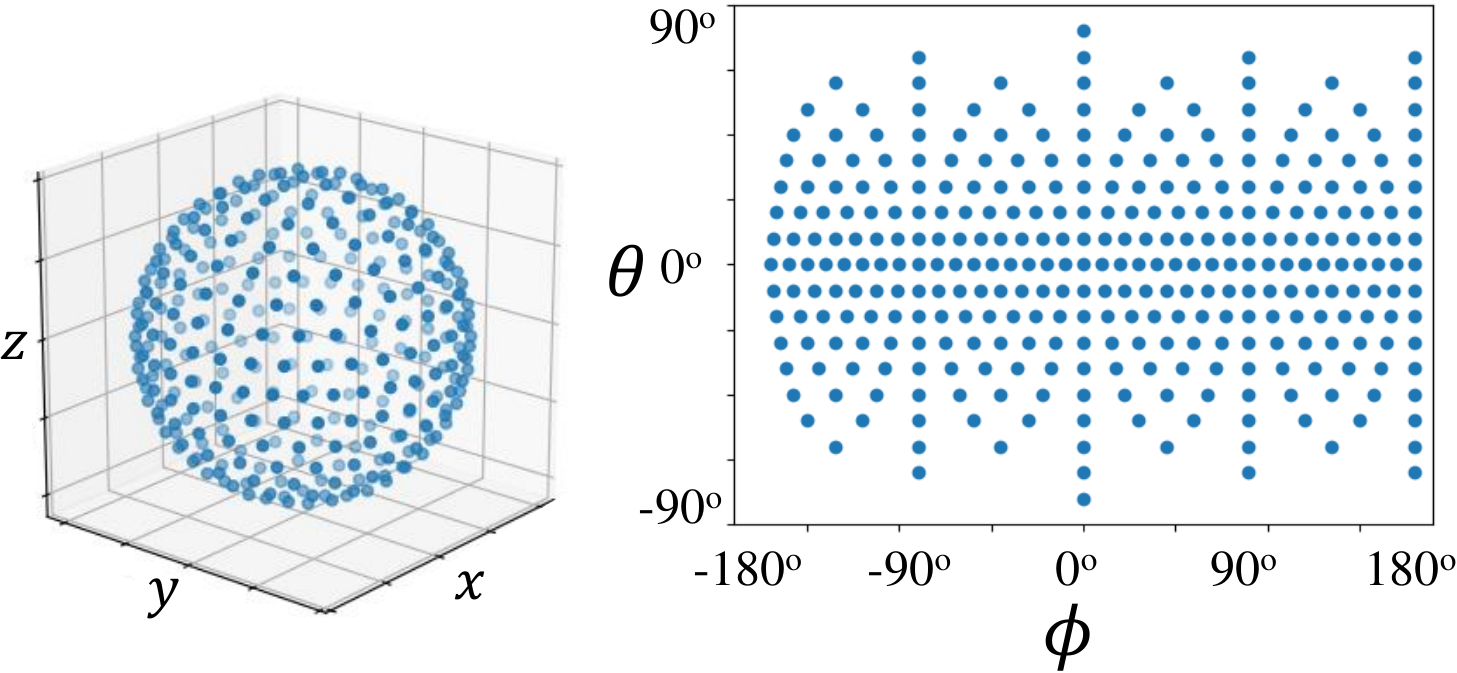}
  \caption{Directions of HRTF measured.}
  \label{fig:hrtf-measure}
\end{figure}

In order to confirm that the developed human mimetic binaural ear unit causes complex frequency response, we measured the HRTF of the unit.
HRTF measurement is conducted in a conference room with the noise level of 35 dBSPL and the reverberation time $RT_{60}$ of 550$\sim$600 msec.
HRTF measurement is taken in 326 directions around the dummy head as shown in \figref{fig:hrtf-measure}, and the angle between each direction and the nearest direction is $10^{\circ} \sim 11.8^{\circ}$.

\begin{figure*}[bt]
  \centering
  \includegraphics[width=1.9\columnwidth]{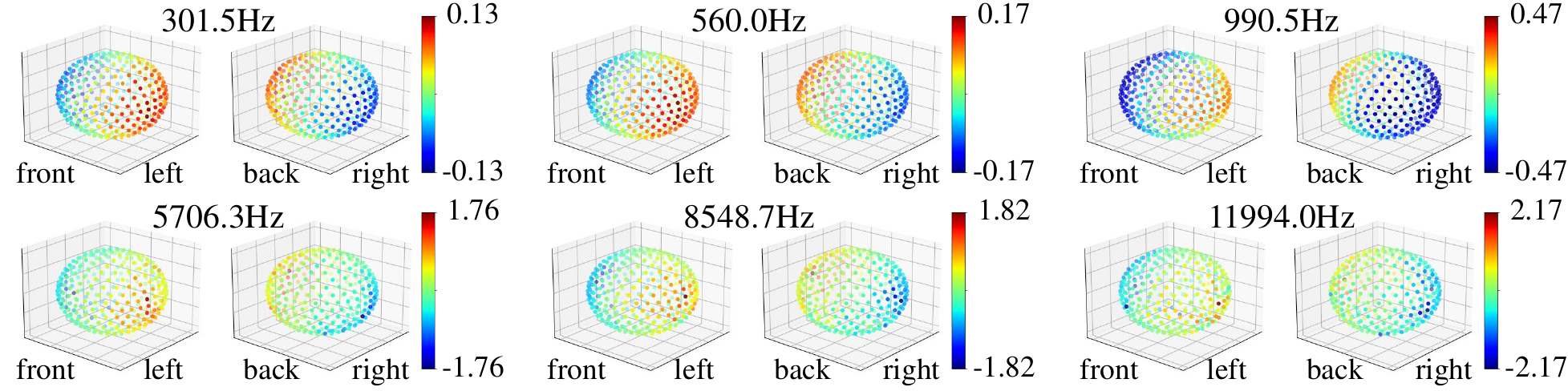}
  \caption{ILDs calculated from HRTFs for each frequency.}
  \label{fig:hrtf-ILD}
\end{figure*}

\begin{figure*}[bt]
  \centering
  \includegraphics[width=1.9\columnwidth]{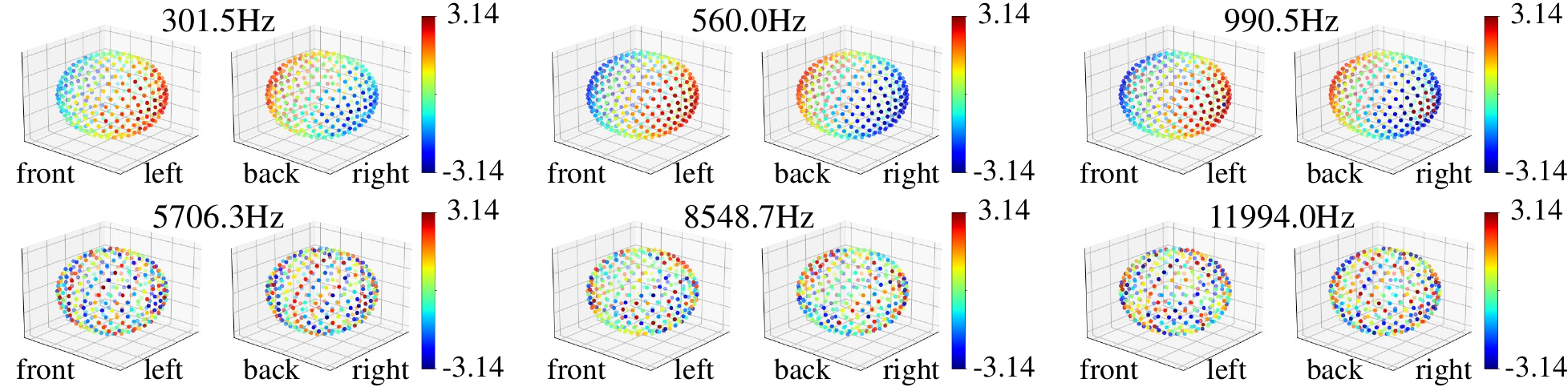}
  \caption{IPDs calculated from HRTFs for each frequency.}
  \label{fig:hrtf-IPD}
\end{figure*}

The results of this experiment are shown in \figref{fig:hrtf-ILD} and \figref{fig:hrtf-IPD}.
ILD and IPD of this experiment are calculated as follows,

\begin{gather}
  ILD_{HRTF}(\omega) = \log |A_{l}(\bm{d}, \omega)| - \log |A_{r}(\bm{d}, \omega)| \\
  IPD_{HRTF}(\omega) = \arg \frac{A_{l}(\bm{d}, \omega)}{A_{r}(\bm{d}, \omega)}
\end{gather}

where $\mathrm{arg}$ is the argument of the complex.

First, we describe the result of ILD.
In lower frequency bands of 301.5, 560.0, and 990.5 Hz, ILD changes gently from left to right, and the range of ILD is relatively small, ranging from 0.13 to 0.47.
On the other hand, in higher frequency bands of 5706.3, 8548.7, 11994.0 Hz, ILD changes in such a way that there are sharp peaks on both sides, and the range of ILD increases from 1.76 to 2.17, indicating that the difference between left and right sides increased.

Next, we describe the result of IPD.
Like ILD results, IPD changes gently from left to right in lower frequency bands.
On the other hand, in higher frequency bands, IPD changes complicatedly because the period of sound is much smaller than the arrival time difference.


\subsection*{Sound source direction estimation using SSDENet}
\subsubsection*{Simulation}

We compare the performance with the MUSIC method \cite{MUSIC} in sound source direction estimation when some types of sounds are generated from virtual sources located in the direction where HRTFs are measured.
In this experiment, a single source is assumed.
In the MUSIC method, the sound source direction is estimated by summing the spatial spectra for detected frequency, and the frequency components used for sound source direction estimation are selected based on the ratio of the eigenvalues of the spatial correlation matrix \cite{mohan2008localization}.
The steering vectors for each frequency in the MUSIC method are the HRTFs measured at each frequency in the previous experiment.
Spatial smoothing is not applied because the size of the correlation matrix is 2 $\times$ 2.
SSDENets for each frequency component are trained by 1000 sounds generated randomly.
The sounds generated by the virtual sound source are sine waves, triangle waves, square waves, and sawtooth waves with fundamental frequencies of 500, 1000, and 2000 Hz, and white noise.
 
\begin{figure*}[bt]
  \centering
  \includegraphics[width=1.9\columnwidth]{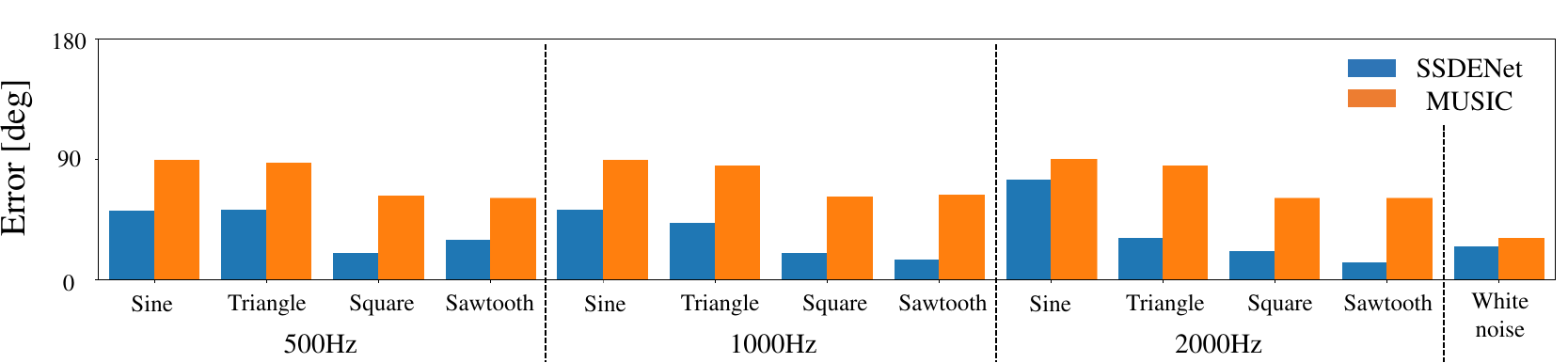}
  \caption{Results of simulation.
  Mean angle error for each condition.}
  \label{fig:sim-result}
\end{figure*}

The results of sound source direction estimation for each wave are shown in \figref{fig:sim-result}.
The mean sound source direction estimation errors of the proposed method are smaller than those of the MUSIC method under all conditions.
In particular, the errors of the MUSIC method for sine waves and triangle waves are around $90^{\circ}$, which is comparable to the random case, but the results of the proposed method are relatively accurate in many directions.

\subsubsection*{Real environment}

We investigated the accuracy of the sound source direction estimation in the real environment using the dummy head shown in \figref{fig:dummy}(a).
This experiment is conducted in a room with the noise level of 58 dBSPL and the reverberation time $RT_{60}$ of 270$\sim$320 msec.

A loudspeaker is placed 1.5 m in front of the dummy head to play white noise of around 73$\sim$76 dBSPL.
We investigate the following three points in this experiment,
\begin{itemize}
  \item sound source direction estimation error in the horizontal plane
  \item sound source direction estimation error in the median plane
  \item discrimination between left and right in the horizontal plane, and between top and bottom in the median plane.
\end{itemize}

\begin{figure*}[tb]
  \centering
  \includegraphics[width=1.9\columnwidth]{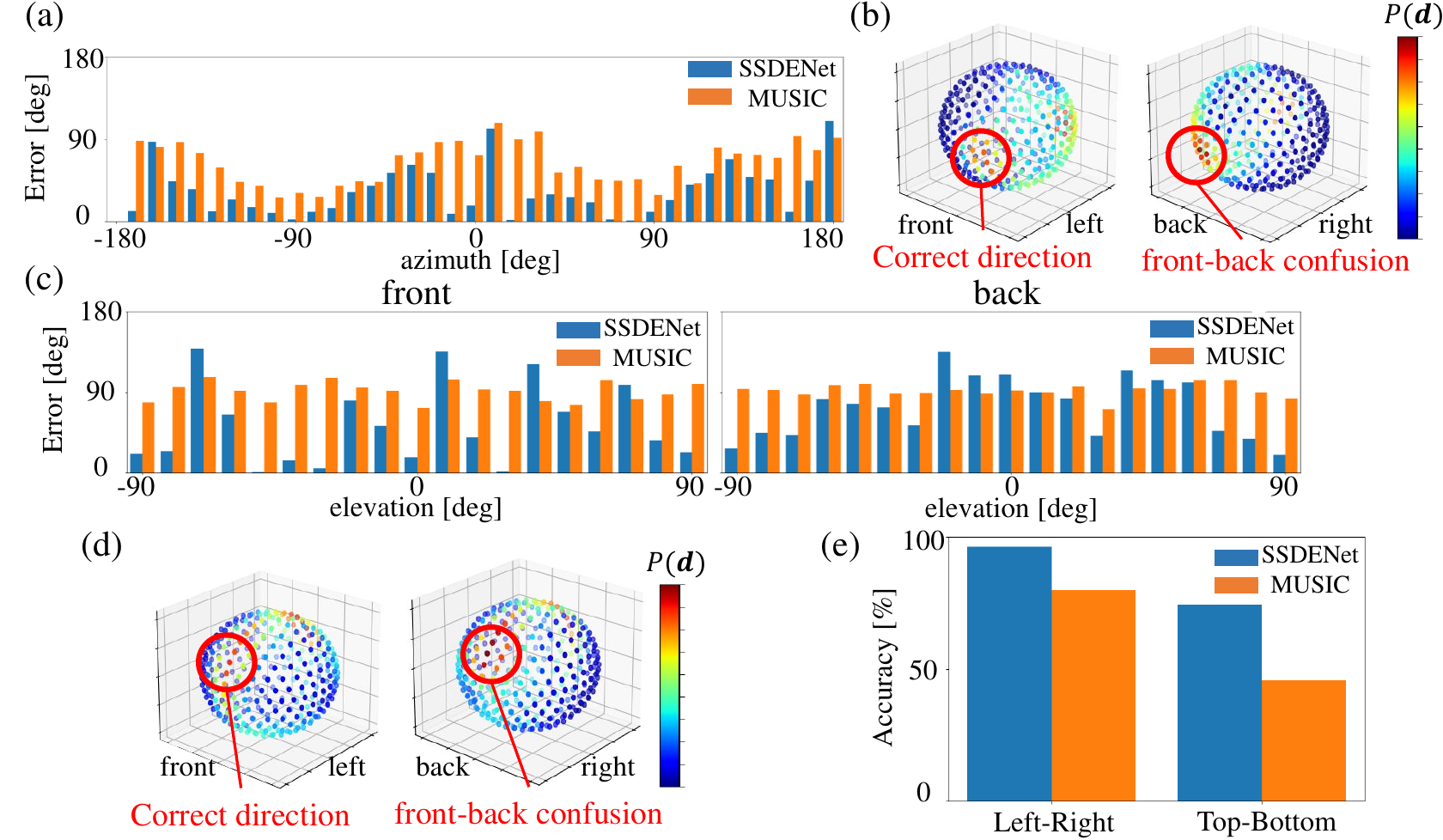}
  \caption{Results of real environment.
  (a) Mean angle error of sound source direction estimation in the horizontal plane.
  (b) Example of estimation result in the horizontal plane. Sound source direction: $\phi = 10^{\circ}, \theta = 0^{\circ}$.
  (c) Mean angle error of sound source direction estimation in the median plane.
  (d) Example of estimation result in the median plane. Sound source direction: $\phi = 0^{\circ}, \theta = 10^{\circ}$.
  (e) (left) Accuracy of Left-Right discrimination in the horizontal plane. (right) Accuracy of Top-Bottom discrimination in the median plane.}
  \label{fig:real-result}
\end{figure*}

The results of this experiment are shown in \figref{fig:real-result}(e).
As shown in \figref{fig:real-result}(a), the error of the proposed method is smaller than those of the MUSIC method in many directions of the horizontal plane.
The results of sound source existences for the direction where the error of the proposed method is large are shown in \figref{fig:real-result}(b).
Although there is a peak in the correct sound source direction, there is also a large peak in the front-back symmetry direction.

In the median plane, the errors of the MUSIC method are around $90^{\circ}$ in any direction, which shows that the MUSIC method is not capable of sound source direction estimation in the median plane.
On the other hand, the proposed method can estimate sound source direction with a sufficiently small error depending on the direction.
The results of sound source existences for the direction where the error of the proposed method is large are shown in \figref{fig:real-result}(d).
As in the case of the horizontal plane, there is a peak in the correct sound source direction, but there is also a large peak in the front-back symmetry direction.

The accuracies of left-right discrimination in the horizontal plane and top-bottom discrimination in the median plane are shown in \figref{fig:real-result}(e).
The proposed method can estimate the left and right of sound source direction with high accuracy.
In top-bottom discrimination, the accuracy of the MUSIC method is around 50 \%, which means that the MUSIC method is not capable of sound source top-bottom discrimination.
On the other hand, the proposed method can discriminate top and bottom though not as good as left and right.

\subsection*{Task realization based on detection of environment including out-of-view}

\begin{figure}[tb]
  \centering
  \includegraphics[width=0.95\columnwidth]{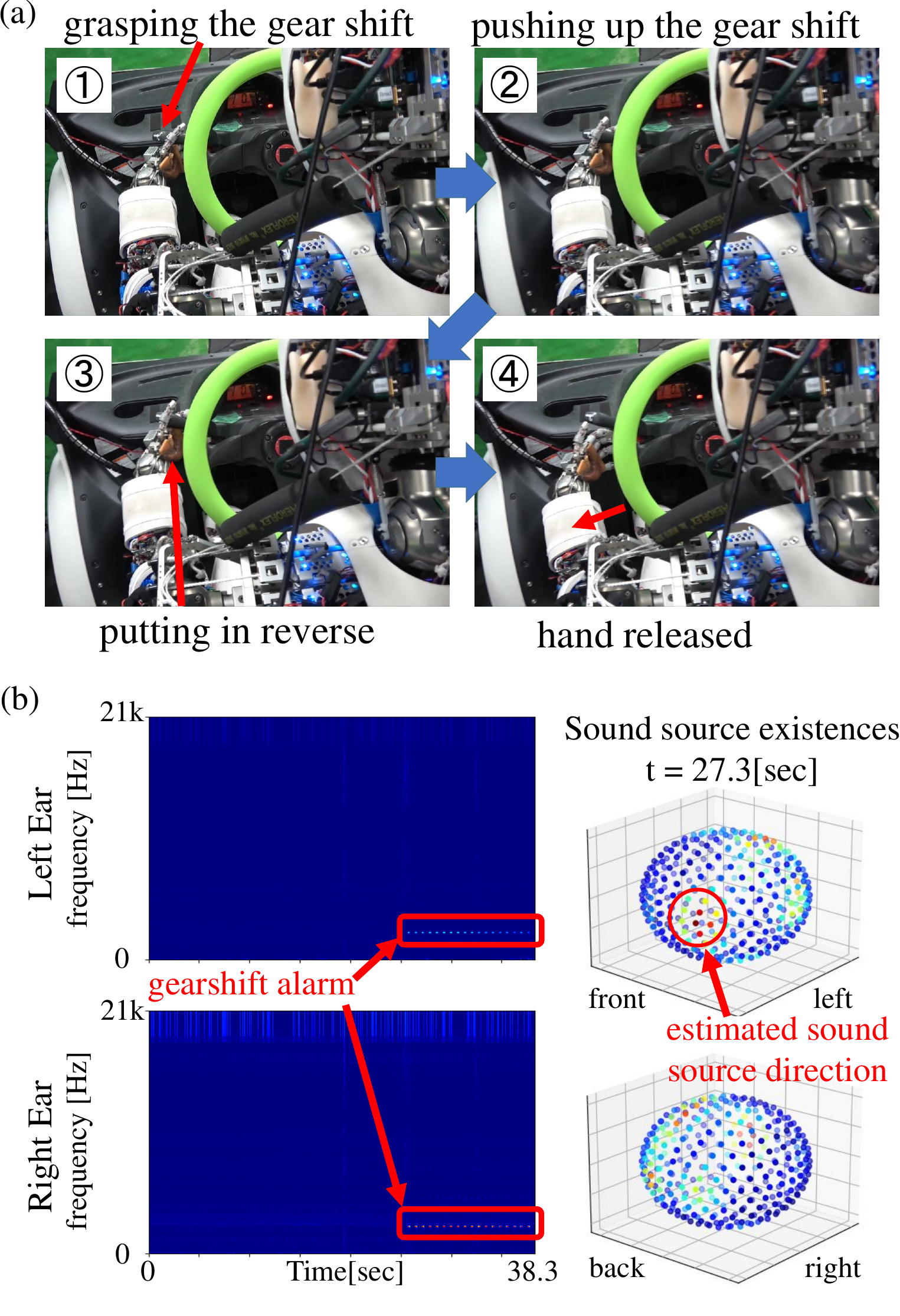}
  \caption{(a) Putting the car into reverse.
  (b) Melspectrograms and result of alarm sound direction estimation.}
  \label{fig:gearshift-exp}
\end{figure}

\begin{figure}[tb]
  \centering
  \includegraphics[width=0.95\columnwidth]{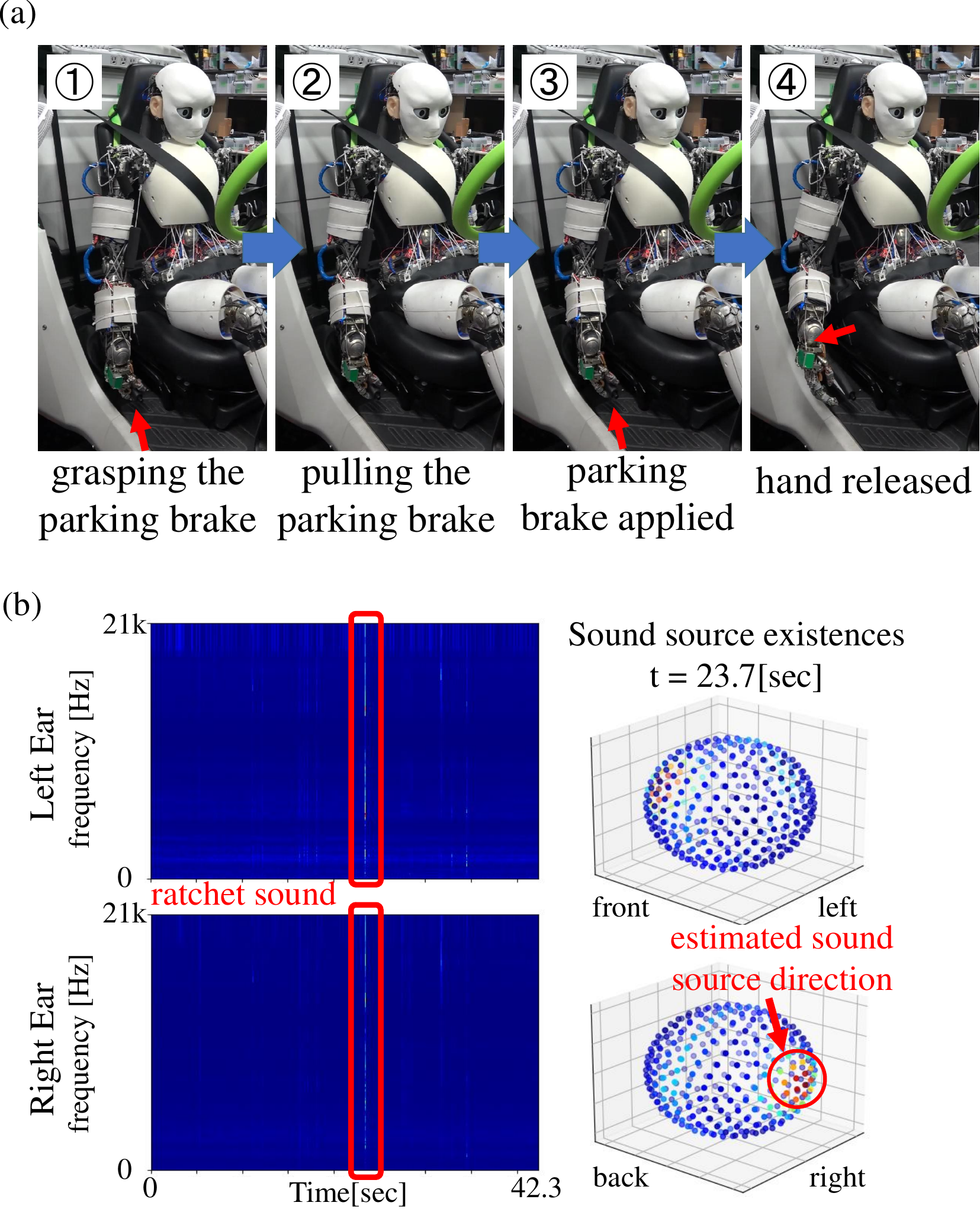}
  \caption{(a) Applying the parking brake.
  (b) Melspectrograms and result of ratchet sound direction estimation.}
  \label{fig:parking-exp}
\end{figure}

\begin{figure}[t]
  \centering
  \includegraphics[width=0.87\columnwidth]{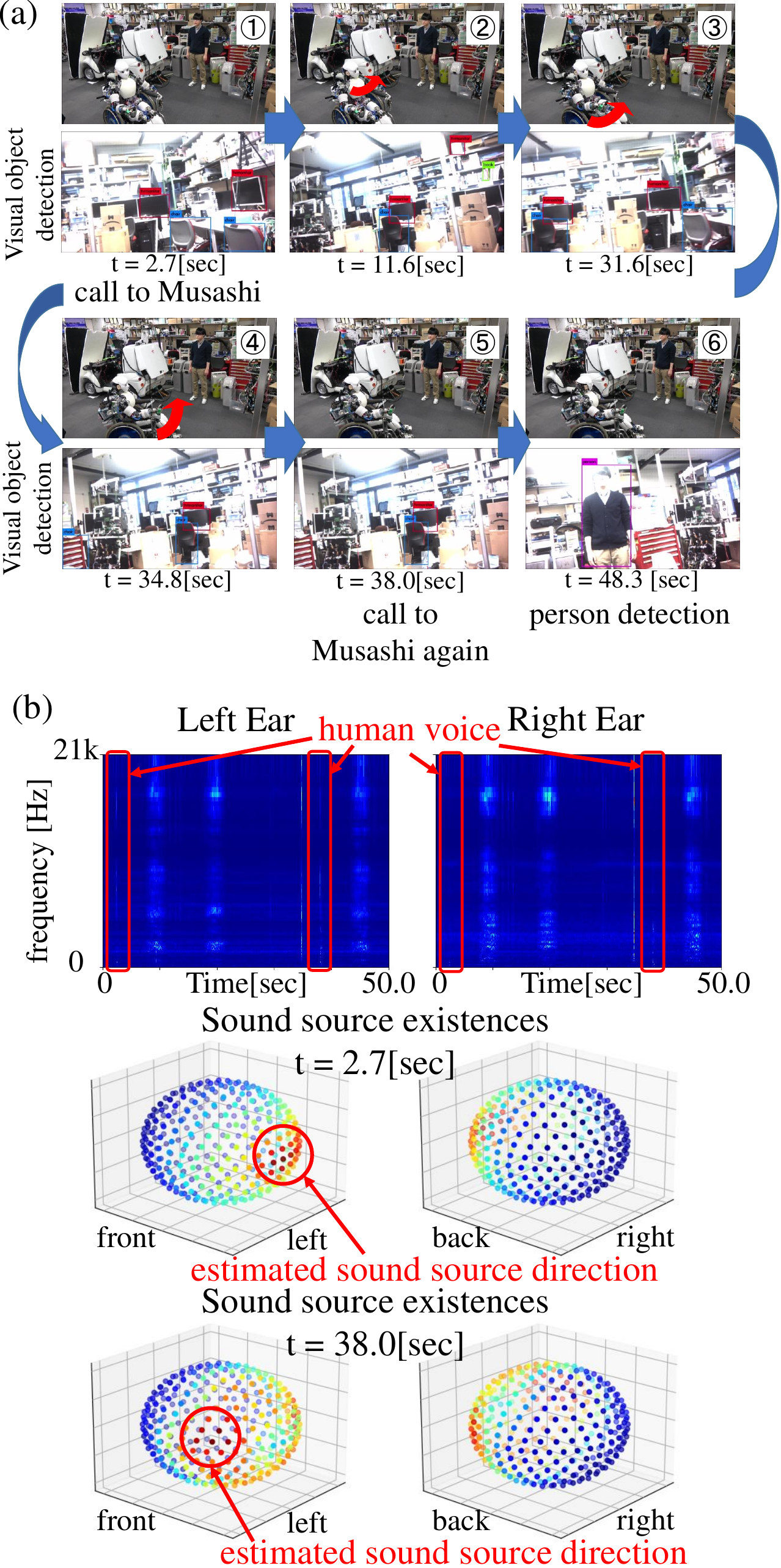}
  \caption{(a) Musashi turns in the direction of the voice.
  (b) Melspectrograms and results of voice direction estimation.}
  \label{fig:wheelchair-exp}
\end{figure}

The musculoskeletal humanoid used in this experiment is Musashi \cite{musashi_iros2019}.
We investigate that the musculoskeletal humanoid can realize tasks that need out-of-view information in the real complex environment.
This experiment contains tasks that require environmental sound detection of not only types of sound but also sound directions.
We use SSDENets trained in the former section.
MelCNN used in this experiment is trained by six types of sound, car horn, alarm, ratchet sound, gearshift manipulating sound, key turning sound, and the human voice.
The 50 samples are recorded for each label.
The F score of MelCNN for these sounds is 0.87.
In this experiment, we deal with the driving noise of Musashi as noise.
We record the driving noise and calculate the driving spectra in advance.
We set the thresholds at 10 times the driving spectra for each frequency and apply the proposed method to spectra exceeding the thresholds.
We use incoming sound and noise as training data for SSDENet and expect Musashi predicts the sound source direction in the real environment with his driving noise.

\subsubsection*{Manipulation of gearshift}

\figref{fig:gearshift-exp} shows the behavior of the gearshift manipulation.
Musashi needs to operate the gearshift in this task and recognize that his operation causes the alarm.
The alarm sound occurs in front of Musashi.
Musashi moves to put the gearshift into reverse and takes the left hand away when Musashi recognizes the alarm sound from the front.
At first, the gearshift is grasped with the left hand as the initial state, and Musashi starts pushing up the gearshift.
At around 25 sec, the gearshift is put into reverse, and the alarm sound starts.
Musashi recognizes the alarm sound direction to be in front at 27.3 sec and releases the left hand from the gearshift to complete the motion.

\subsubsection*{Manipulation of parking brake}

\figref{fig:parking-exp} shows the behavior of the parking brake manipulation.
Musashi needs to operate the parking brake in this task and recognize that his operation causes the ratchet sound.
The ratchet sound occurs right side of Musashi.
Musashi pulls up the parking brake and takes the right hand away when Musashi recognizes the ratchet sound from right.
At first, Musashi grasps the parking brake by the right hand as the initial state and starts pulling up the parking brake.
At around 23.5 sec, the parking brake is applied, and the ratchet sound is heard.
Musashi recognizes the ratchet sound direction to be in the right at 23.7 sec and releases the right hand from the parking brake to complete the motion.

\subsubsection*{Response to calls}

\figref{fig:wheelchair-exp} shows the behavior of the response to call.
In order to turn to the direction of call, Musashi moves the neck and eyes and manipulates the wheelchair to turn the body around, and Musashi captures the person who calls Musashi in the view.
The visual detection of humans is performed for the right eye view, and \cite{yolov3} is used for the detection.
Initially, at around 2 sec, a person calls from the left side of Musashi, and Musashi detects the voice and recognizes that the call comes from the left side.
After that, Musashi moves the neck and eyes at around 11 sec to check the left side, but Musashi cannot catch the person in the field of view, and then Musashi turns the body to the left side by manipulating the wheelchair for around 30 sec.
The person calls again at around 38 sec, and Musashi recognizes that the voice came from the left front.
Musashi moves the neck and eyes to check the left side again, and Musashi captures the person in its field of view and finishes the behavior of the response to call at around 48 sec.

\section*{\discussion} \label{sec:discussion}

We discuss the results obtained from the experiments of this study.
First, we describe the effects of the human mimetic outer ear structure.
The human-like structures of the pinna and ear canal produce changes of frequency response in the median plane depending on the elevation and complex changes of HRTF.
The frequency response changes in the median plane due to the pinna shape show effect similar to the pinna notch in humans.
The pinna notches are seen from about 5000 Hz in the lower frequency range and are clearly visible in the higher frequency range, which corresponds to the previous investigation \cite{musicant1984influence}.
Also, in terms of the effect on the HRTF in the frequency domain, at lower frequencies, ILD changes little, and the effect by ITD (IPD) is large, while the effect by ILD increases and ITD (IPD) has complex changes at higher frequencies, which supports "Duplex Theory" \cite{rayleigh1907}.
The human mimetic binaural ear unit developed in this study reproduces the characteristics of the human outer ear.

Second, we describe the sound source direction estimation using SSDENet.
In the simulation with multiple types of sound sources, the estimation errors are smaller than those of the MUSIC method.
In the existing sound source direction estimation methods such as the MUSIC method, each frequency and direction is calculated independently.
On the other hand, SSDENet has a structure that includes the relationship between directions to be estimated in the proposed method so that it is expected for the false estimation results to be suppressed.
Also, SSDENet works robustly in the real environment because it uses data with background noise components in training.
Both the MUSIC method and the proposed method using SSDENet show higher accuracy for sounds with more frequency components such as square waves, sawtooth waves, and white noise than for sounds with fewer frequency components such as sine waves and triangle waves.
The reason is that the effect of the phase diversity of IPD becomes smaller as the number of frequency components increases.
In the real environment, the accuracy of the proposed method is higher than that of the MUSIC method in the horizontal plane.
Although the errors in the median plane are not as large as that of the MUSIC plane in any direction, the estimation errors are larger than those of the horizontal plane, and it can be said that it is only for reference in practical use.
Also, there are front-back confusions in both the horizontal and median planes.
The accuracy of human sound source direction estimation is lower for top-bottom and front-back directions than for left-right directions.
It can be said that the proposed method shows a similar trend of accuracy in sound source direction estimation as humans.
However, humans improve their sound source localization accuracy by rotating and tilting their heads.
In order to realize human-like sound source direction estimation, recognition combined with motion should be addressed in future work.

Finally, we describe task realization based on environment recognition, including out-of-view.
We realize the acquisition of environment information and task realization based on human-like auditory information processing for each action.
The high accuracy of the proposed system enables auditory recognition even in a cluttered real environment with background noise.

\section*{\conclusion} \label{sec:conclusion}

In this research, we proposed a human mimetic auditory environmental recognition system consisting of a human mimetic binaural ear, sound source direction estimation system, and environmental sound detection system.
The developed human mimetic binaural ear unit, which consists of the human mimetic outer ear structure, the microphone board that mimics human hearing characteristics, and the acoustic processing board that performs frequency decomposition with low latency, shows complex frequency response depending on the sound source direction like the human ear.
The proposed sound source direction estimation method mimics detection of ILD and ITD at SOC and response to each direction at IC in human auditory information processing.
In contrast to the existing engineering method, the proposed method uses neural networks that include the relationship between directions and shows that it is possible to estimate the sound source direction using only two ears roughly.
By implementing the proposed system, we realized for the musculoskeletal humanoid to recognize the environment, including out-of-view areas, and perform some tasks which require recognition of objects out of view.
In future works, we will work on the human mimetic wide-range environment recognition and task realization by integrating the proposed method with sensing modes, such as visual and tactile senses.

\begin{backmatter}
\section*{Acknowledgements}
Not applicable

\section*{Authors' contributions}
YO and KK proposed the concept of the human mimetic auditory information processing system.
YO and YN developed the human mimetic ear unit.
YO, KK, YK, MN and YT supported experiments of this research.
YA, KO, KK and MI supported the whole development of this research.
All authors read and approved this manuscript.

\section*{Funding}
This research was partially supported by JST ACT-X Grant JPMJAX20A5.

\section*{Availability of data and materials}
Not applicable
\end{backmatter}

\section*{Declarations}

%

\begin{backmatter}
\section*{Competing interests}
The authors declare that they have no competing interests.

\bibliographystyle{bmc-mathphys}
\bibliography{main}

\end{backmatter}
\end{document}